\documentclass[letterpaper, 10 pt, conference]{ieeeconf}  

\IEEEoverridecommandlockouts                              

\usepackage{color}
\usepackage{lipsum}
\usepackage{amsmath} 
\usepackage{algorithmic} 
\usepackage{algorithm} 
\usepackage{siunitx} 
\usepackage{anyfontsize} 
\usepackage{t1enc} 
\usepackage{multirow}  
\usepackage{amssymb}
\usepackage{pifont}
\usepackage{graphicx}
\usepackage{subcaption}
\usepackage{booktabs} 

\title{\LARGE \bf
DMSA - Dense Multi Scan Adjustment for LiDAR Inertial Odometry and Global Optimization
}

\author{David Skuddis$^{1}$ and Norbert Haala$^{1}$
\thanks{$^{1}$Both authors are with Institute for Photogrammetry and Geoinformatics, University of Stuttgart, 70174 Stuttgart, Germany
        {\tt\footnotesize \{david.skuddis,\,norbert.haala\}@ifp.uni-stuttgart.de}}%
}

\begin{document}

\maketitle
\thispagestyle{empty}
\pagestyle{empty}

\begin{abstract}
We propose a new method for fine registering multiple point clouds simultaneously.
The approach is characterized by being dense, therefore point clouds are not reduced to pre-selected features in advance. Furthermore, the approach is robust against small overlaps and dynamic objects, since no direct correspondences are assumed between point clouds.
Instead, all points are merged into a global point cloud, whose scattering is then iteratively reduced. This is achieved by dividing the global point cloud into uniform grid cells whose contents are subsequently modeled by normal distributions.
We show that the proposed approach can be used in a sliding window continuous trajectory optimization combined with IMU measurements to obtain a highly accurate and robust LiDAR inertial odometry estimation.
Furthermore, we show that the proposed approach is also suitable for large scale keyframe optimization to increase accuracy. We provide the source code and some experimental data on https://github.com/davidskdds/DMSA\_LiDAR\_SLAM.git.

\end{abstract}
\section{INTRODUCTION}
The field of point cloud registration and simultaneous localization and mapping (SLAM) has been researched for several decades and there are already a large amount of mature approaches and implementations. The task here is to estimate the sensors trajectory using point clouds (and IMU data). Current challenges in this area are, in particular, achieving high accuracies (mm-range) and increasing robustness. In terms of robustness, especially narrow spaces, highly dynamic motion and dynamic objects are current challenges. In this paper, we present a new approach that aims to take us one step closer to solving these challenges.
\newline\newline
The main contribution of this work is the proposal of a new method for dense multi scan adjustment. While other methods for point cloud registration already use multivariate normal distributions to separately model point clouds within a registration process e.g. \cite{biber2003normal},\,\cite{magnusson2009three},\,\cite{myronenko2010point},\,\cite{yokozuka2021litamin2}, our method considers the points from two or more point clouds in a common frame as one set of multivariate normal distributions. This approach has two main advantages: the normal distributions updated in each iteration serve as weights in the optimization, which is why non-matching shapes from e.g. dynamic objects are automatically weighted lower. An outlier rejection mechanism is therefore not necessary.
The second advantage is that there are no direct correspondences between point clouds to be registered. Instead, there are only correspondences from each point to its respective normal distribution, thus number of correspondences is largely reduced. As a result, multiple point clouds can be optimized efficiently at the same time.
Furthermore, we have developed a LiDAR inertial odometry algorithm based the method that performs superior to other state-of-the-art methods in terms of robustness as well as accuracy in our experiments with a challenging selection of data sequences.
\begin{figure}[ht!]
\begin{center}
		\includegraphics[width=0.95\columnwidth,trim=0 20 0 70,clip]{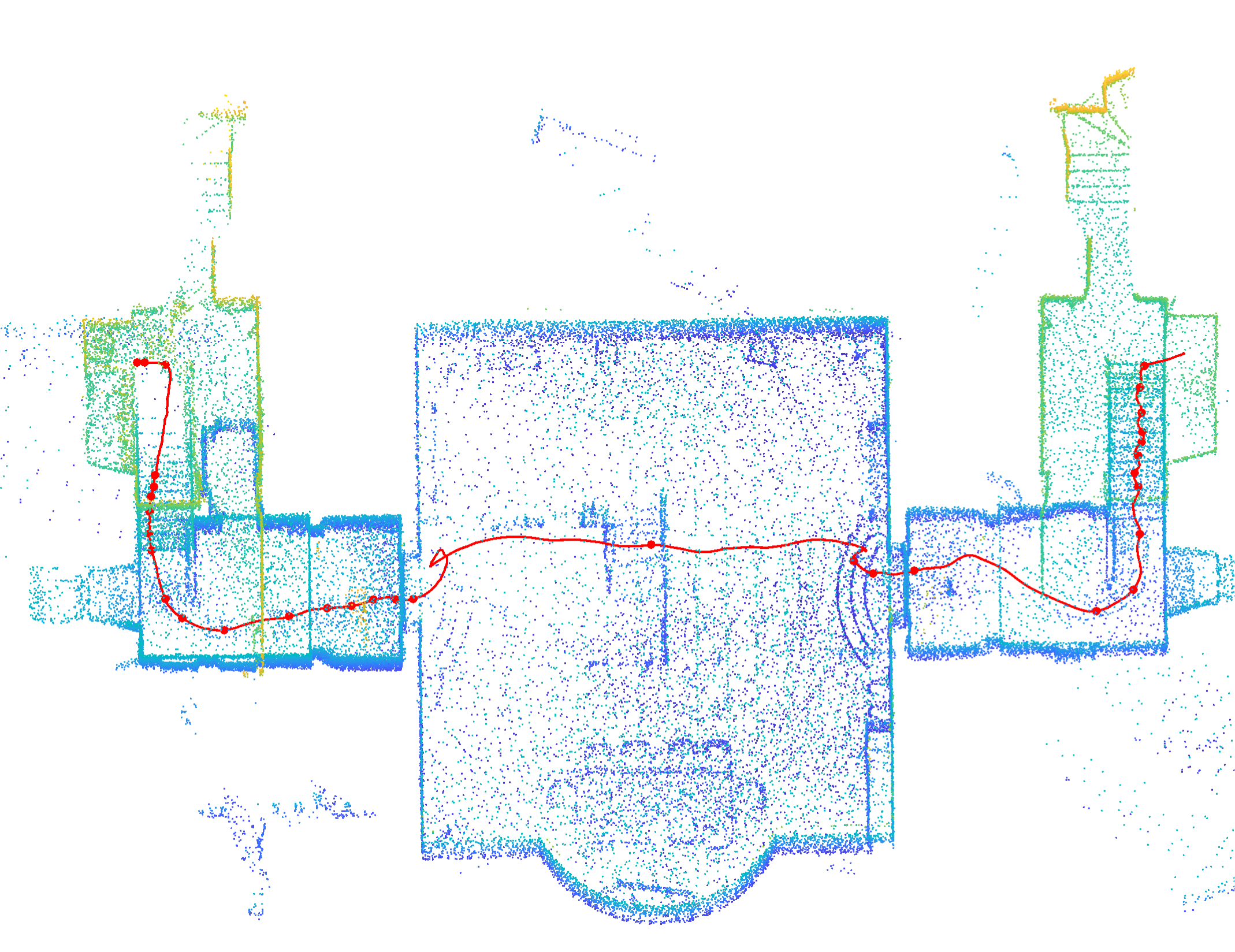}
	\caption{Resulting sparse keyframe point cloud of sequence \textit{exp14 Basement} of the Hilti-Oxford Dataset \cite{9968057}. The estimated trajectory is marked with a red line and keyframe positions are outlined with red dots.}
\label{fig:basement_bv}
\end{center}
\end{figure}
\begin{figure*}[htbp]
\begin{center}
		\includegraphics[width=1.8\columnwidth]{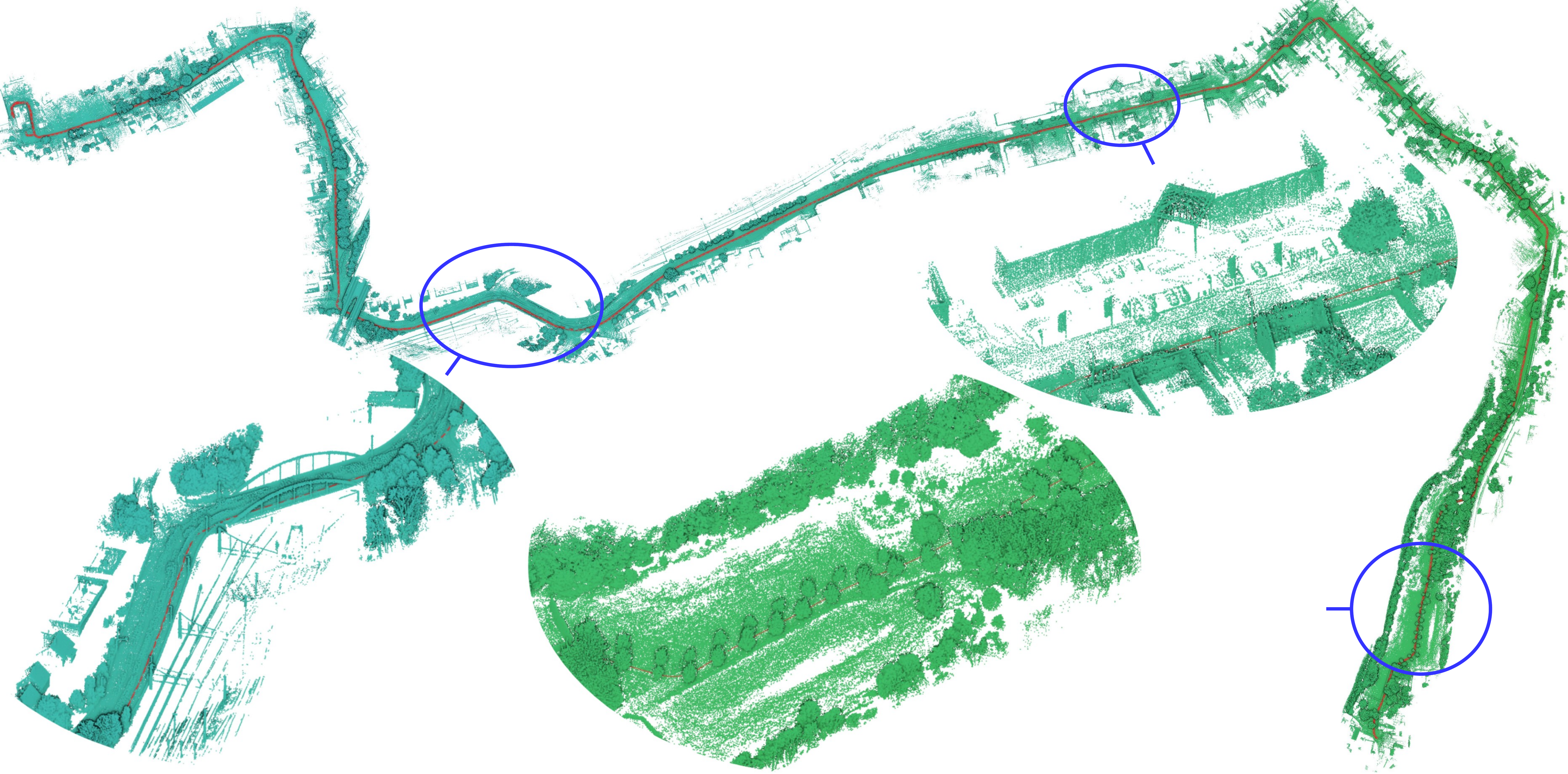}
	\caption{Resulting keyframe point cloud from sequence \textit{Bicycle Street} of our data recording. The scene contains urban areas, field paths and dynamic objects. The trajectory of the sensor platform mounted on a bicycle trailer is marked in red. For areas marked in blue, detail views are provided.}
\label{fig:stitch_with_circles}
\end{center}
\end{figure*}
\section{Related Work}
In this chapter we give a brief overview about point cloud registration, LiDAR (inertial) odometry and multi scan registration methods for global optimization.
\subsection{Point Cloud Registration}\label{chap::pcreg}
The field of point cloud registration is very broad and there is an extensive number of methods. In our literature review, we restrict attention only to convergence-based approaches. Probably the most famous algorithm in the field of point cloud registration is the ICP algorithm \cite{besl1992method}, from which many other approaches are derived \cite{low2004linear},\,\cite{segal2009generalized}. There are still modern SLAM systems that are based on the ICP algorithm \cite{vizzo2023ral}. 
Another group of methods for point cloud registration transforms one of the two point clouds into sets of normal distributions and then optimizes the relative pose of the second point cloud by maximizing the likelihood, in some works represented as a score, of the points of the second point cloud \cite{biber2003normal},\,\cite{magnusson2009three},\,\cite{myronenko2010point}.
There are also concepts, where both point clouds to be registered are converted to normal distributions and the error between the Normal distributions is minimized using metrics like KL divergence \cite{yokozuka2021litamin2}.
All these approaches assume correspondences between nearest points or nearest normal distributions between the point clouds to be registered. This can result in erroneous correspondences due to small overlaps or dynamic objects, which reduce the accuracy of the solution found. One way to deal with this is an advanced weighting of correspondences during optimization to reduce the influence of outliers \cite{chebrolu2021ral},\,\cite{vizzo2023ral}.
Other methods avoid the formation of explicit correspondences by minimizing a cost function that is independent of correspondences for the determination of relative poses. \cite{hillemann2019automatic} proposes a correspondenceless method that utilizes eigenvalue-based geometric features for automatic extrinsic calibration. Although extrinsic calibration is a different type of problem, this work inspired us to develop the proposed approach.

\subsection{LiDAR (Inertial) Odometry}
LiDAR Odometry deals explicitly with the application and adaptation of point cloud registration methods to data from mobile laser scanners in order to estimate the trajectory of the sensor. LiDAR \textit{Inertial} Odometry describes corresponding LiDAR odometry methods that additionally include IMU data to support the estimation. 
One of the first robust and accurate LiDAR odometry methods with real-time capability was LOAM \cite{zhang2014loam}. Here, plane and edge features are used to estimate the trajectory. There are various extensions to this method, e.g. for planar moving robots \cite{shan2018lego} or for higher processing speeds \cite{wang2021f}. Other LiDAR odometry methods use surface elements (surfels) as features \cite{behley2018efficient}, which can represent a point cloud relatively densely in most cases.
While these approaches reduce the point clouds to a set of features, there are also dense methods that do not rely on the presence of specific shapes. Examples are CT-ICP \cite{dellenbach2022ct} and KISS-ICP \cite{vizzo2023ral}.
The main advantages of LiDAR inertial odometry, enabled by IMU measurements, are higher robustness in fast motions, stabilization in feature poor locations and the observability of the gravitational direction to stabilize the drift of orientation. Wildcat SLAM \cite{ramezani2022wildcat} extracts surfels from the point clouds and optimizes a continuous trajectory defined by control poses and spline interpolation in a sliding window manner together with the IMU data over several seconds to obtain highly accurate submaps. For each submap, a gravitation direction is determined and submaps are registered relative to each other.
LIO-SAM \cite{shan2020lio} extracts edge and planar features from the point clouds and performs a factor graph optimization along with the IMU data. LiLi-OM \cite{liliom}, which supports both spinning LiDARs and solid-state LiDARs with comparatively small fields of view, also extracts planar and edge features and performs scan-to-map matching. Keyframes are optimized both in a sliding window manner and after loop closure detections including IMU data. FAST-LIO2 \cite{xu2022fast} and Point-LIO \cite{he2023point} perform scan-to-map matching with the LiDAR point clouds and use Kalman filters for joint state estimation with the IMU data.

\subsection{Multi Scan Registration for Global Optimization}
When registering multiple point clouds, the goal is to specifically model and minimize the error between multiple point clouds to obtain a consistent global point cloud. Of course, the previously mentioned procedures for point cloud registration (compare \ref{chap::pcreg}) can also be applied sequentially to point cloud pairs of a larger scene to obtain a merged point cloud. However, this approach can lead to inconsistencies and is not suitable for finding the optimal solution in many cases.
Most of the LiDAR (inertial) odometry methods mentioned above build incremental maps the current scan is aligned to. While some map the environment without optimizing the consistency within the map \cite{xu2022fast}, others perform feature-based optimization only in a sliding time window \cite{zhang2014loam},\,\cite{ramezani2022wildcat}. After loop closure detections, a pose graph optimization is usually performed. In a pose graph optimization, the poses of keyframes are typically optimized on the basis of pairwise alignments. An early work that uses this principle is \cite{lu1997globally}. While these types of procedures are very efficient and improve consistency it is also an indirect concept without specifically modeling global consistency.
An intuitive approach for multi scan registration is to define an error function that represents the correspondences between all point cloud pairs with overlap and then minimize them. While previous works have studied this approach \cite{benjemaa1998solution}, it is computationally expensive, since the possible number of correspondences grows quadratically with the number of keyframes.
An early approach uses range image to optimize a set of planar points from different scans \cite{chenobject}. Other works extract planes from point clouds and perform global optimization based on the planes within the SLAM context \cite{kaess2015simultaneous},\,\cite{hsiao2017keyframe}. In \cite{liu2021balm} BALM is presented, an efficient analytical solution of an adapted bundle adjustment for LiDAR based on plane and edge features.
Using features such as planes and edges for keyframe optimization can, as shown in \cite{liu2021balm}, improve the accuracy of the resulting trajectory as well as global consistency in many cases. At the same time, when reducing the point clouds to a set of features, information may be lost and there may be corner cases where corresponding features are not available.
With our work we present a method that can optimize dense keyframes without a reduction to features. 
In addition, our approach is efficient because no direct correspondences need to be formed between all the point sets. 
\section{Methodology}
In this chapter, we describe the proposed approach in detail.
First, we explain the basic algorithm. Then, we explain how the DMSA approach can be combined with IMU measurements to obtain accurate and robust LiDAR inertial odometry. Subsequently, we show how the approach can be further used to optimize keyframes, both after loop closures and in a sliding window manner to further increase accuracy.
\begin{figure}[h!]
\begin{center}
		\includegraphics[width=0.9\columnwidth]{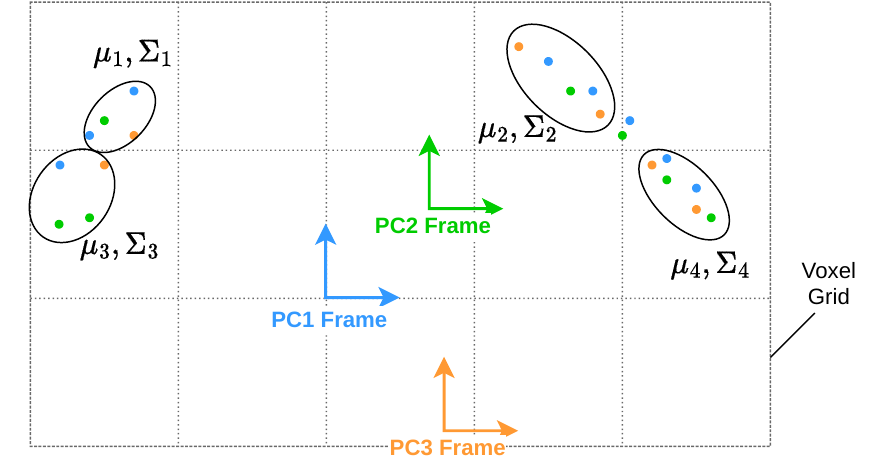}
	\caption{Exemplary two-dimensional setup with three point clouds. In this case, point cloud 1 (PC1, blue) is used as reference and the relative transformations of point cloud 2 and 3 (PC2 and PC3) to point cloud 1 are optimized. Four voxels contain enough points to be considered as landmark.}
\label{fig:dmsa_example_setup}
\end{center}
\end{figure}

\subsection{Dense Multi Scan Adjustment (DMSA)} \label{chap:dmsa_core}
The basic idea behind the DMSA algorithm is a landmark-based graph optimization. Instead of using pre-selected landmark types like planes and edges, the environment is modeled with multivariate normal distributions.
More precisely, the space is divided into voxels hierarchically with two levels. For better illustration, only one grid is shown in each of the following sketches. Each voxel containing more than $n_{min}$ points is considered a landmark. It should be noted that due to the two different sized voxel grids, a single LiDAR point may be represented in two landmarks. In the following we call a set of points, that form a landmark, $L_j$ and the number of points of a landmark $\mathrm{n}(L_j)$. The position of the $j^{th}$ landmark $L_j$ is defined by the mean $\mu_j$ of all points in the cell and the shape distribution is described by the points covariance $\Sigma_j$. An exemplary setup is outlined in Fig. \ref{fig:dmsa_example_setup}.\newline
The target is now to minimize the distances of all measured points $p_{L_j,k}$ to the respective landmark means $\mu_j$ of their cells, taking into account the landmark shape distributions. The approach can be summarized with formula \ref{equ:mainFormula}:
\begin{equation}\label{equ:mainFormula}
	\min_{} \{ \sum\limits_{j=1}^{n_L} \frac{1}{\mathrm{n}(L_j)} \sum\limits_{k=1}^{\mathrm{n}(L_j)}(p_{L_j,k}-\mu_j)^\intercal\Sigma^{-1}_j(p_{L_j,k}-\mu_j) \}
\end{equation}
Weighting with the inverse number of points per landmark prevents overfitting of locations with high point densities.
At this point, the coupling of the error weighting by the inverse covariance matrix $\Sigma^{-1}_j$ with the errors to be minimized should be noted. The smaller the distances of the points $p_{L_j,k}$ to the center of the respective landmarks $\mu_j$ become, the stronger the errors are subsequently weighted.
The described error function is minimized iteratively using Levenberg-Marquardt \cite{more2006levenberg} optimization steps.
During each optimization step, the covariance matrices of the landmarks are assumed to be constant and the errors of the points to their respective landmarks are minimized. Here landmarks are not distinguished between inliers and outliers. 
\begin{algorithm}[h!]
 \caption{The DMSA Algorithm}\label{alg::core}
 \begin{algorithmic}[1]
 \renewcommand{\algorithmicrequire}{\textbf{Input:}}
 \renewcommand{\algorithmicensure}{\textbf{Output:}}

  \REQUIRE $n_P$ point clouds $P_{s,i}$ in sensor frame with initial transforms set $\mathbf{T^w_{s,\mathrm{init}}}=(T^w_{s,1},...,T^w_{s,n_P})$ with $T^w_{s,i}= (r_1,r_2,r_3,x,y,z),i\in \{1,\,...\,,n_P\}$ from sensor frame (index $s$) to reference frame (index $w$)
\STATE $\mathbf{T^w_{s}}=\mathbf{T^w_{s,\mathrm{init}}}$
  \WHILE{not converged}
  \STATE Transform point clouds to a common reference frame
  \STATE Split all points in the reference frame into voxels twice, once with a coarse grid size and once with a fine grid size.
  \STATE Remove voxels that contain less than $n_{min}$ points
  \STATE Init error vector $E\in {\rm I\!R}^{\mathrm{n}(L_j)\times 1}$
  
  \FORALL{voxels with their respective point sets $L_j$}
  \STATE Calculate mean $\mu_j$ and covariance $\Sigma_j$ of points
  \STATE Calculate weight $w_j=\frac{1}{\mathrm{n}(L_j)}$

  \FORALL{points $p_k,k\in \{1,\,...\,,\mathrm{n}(L_j)\}$ in $L_j$}
   \STATE $E(j)+=w_j(p_k-\mu_j)\Sigma^{-1}_j(p_k-\mu_j)$
     
  \ENDFOR
  \ENDFOR
  \STATE Calculate numeric Jacobian $\mathbf{J}_{E}(\mathbf{T^w_{s}})$ while keeping the covariances unchanged
  \STATE Update $\mathbf{T^w_{s}}$ with Levenberg–Marquardt algorithm \cite{more2006levenberg}
  \ENDWHILE

 \RETURN $\mathbf{T^w_{s}}$
 \end{algorithmic}
 \end{algorithm}

 \begin{figure}[h!]
\begin{center}
		\includegraphics[width=0.90\columnwidth]{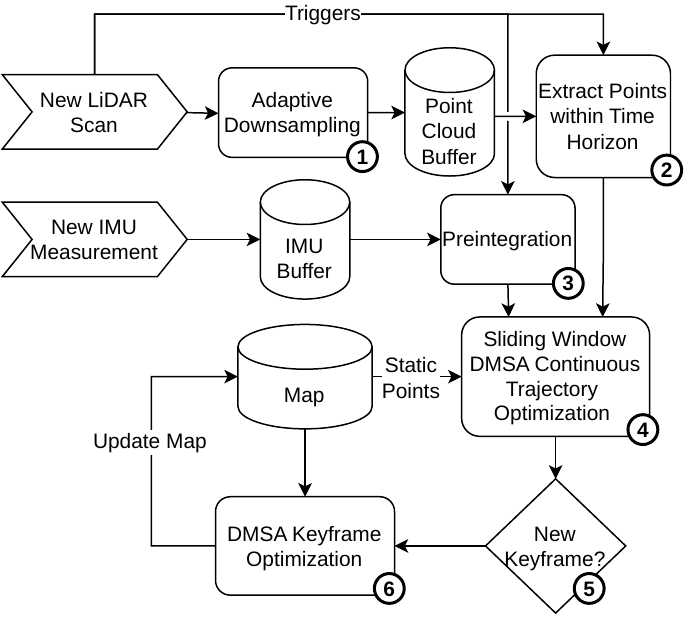}
	\caption{DMSA LiDAR Inertial Odometry overview. Numbers indicate the processing order.}
\label{fig:dmsa_lio_float}
\end{center}
\end{figure}
\subsection{DMSA LiDAR Inertial Odometry}
We have developed a LiDAR inertial odometry system based on the previously described DMSA algorithm. Fig. \ref{fig:dmsa_lio_float} gives an overview of the structure of the approach.
New scans are first adaptively downsampled (see section \ref{chap:downsampling}) and then stored in a ring buffer in the same manner as IMU measurements. After the arrival of a new scan, all points within a sliding time window are extracted from the ring buffer. The trajectory is parameterized by control poses within the time window. To support the subsequent optimization, IMU measurements are preintegrated as proposed in \cite{forster2016manifold} between two control poses. Then, a continuous trajectory is optimized based on the points within the time window, the preintegrated IMU measurements, and static points derived from the map. This process is the core of DMSA LiDAR Inertial Odometry and is described in more detail in section \ref{chap:sliding_window_opt}.
\subsubsection{Adaptive Downsampling} \label{chap:downsampling}
To reduce the number of points we use a grid random filter. Here, the point cloud is divided into uniform grid cells. Then one point is randomly selected from each grid cell. The downsampled point cloud is the set of all selected points and the filtered point count corresponds to the set of grid cells containing points. We decided to use an adaptive grid size to be able to use the system in different environments without adjustments.
The system automatically switches between four different resolution levels. The goal is to reach a minimum number of points $N_{min\_pts}$.
The system starts with the coarsest resolution level. If the number of points after downsampling is smaller than $N_{min\_pts}$, the next finer resolution is used. If the number of points greater or equal to $N_{min\_pts}$, the corresponding resolution is used. After downsampling, the most distant points are step by step removed until the number of points $N_{min\_pts}$ is reached or the current point falls below a distance threshold.
\begin{figure}[h!]
\begin{center}
		\includegraphics[width=0.8\columnwidth,trim=0 40 0 0,clip]{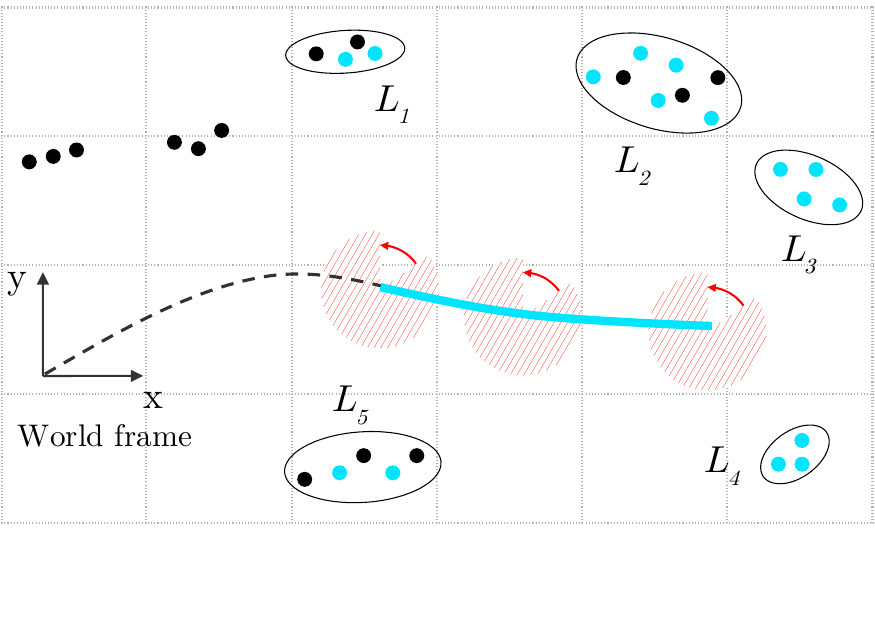}
	\caption{Outline of the sliding window continuous trajectory optimization. The teal line represents the continuous trajectory within the sliding time window to be optimized. Teal dots symbolize LiDAR points acquired within the time window. Black dots represent \textit{Static Points} from the map. The points belonging to the landmarks $L_1$ to $L_5$ are marked with black ellipses.}
\label{fig:sliding_window_sketch}
\end{center}
\end{figure}
\begin{table*}[htbp!]
\centering
\caption{Results of our experiments: APE w.r.t. translation part in meters (with SE(3) Umeyama alignment) using \cite{grupp2017evo} of selected sequences from the Hilti-Oxford Dataset \cite{9968057}, the Newer College Dataset \cite{ramezani2020newer} and own data recordings. The lowest rmse value and lowest maximum error of each sequence are bold. All procedures were performed with the same functional parameter settings for all sequences. Each approach had identical functional parameter settings for all sequences.}
\resizebox{0.9\textwidth}{!}{
\begin{tabular}{@{}llcccccccccc@{}}
\toprule
\multicolumn{1}{c}{\multirow{2}{*}{Dataset}} & \multicolumn{1}{c}{\multirow{2}{*}{Sequence}} & \multirow{2}{*}{\begin{tabular}[c]{@{}c@{}}Duration\\ {[}s{]}\end{tabular}} & \multirow{2}{*}{\begin{tabular}[c]{@{}c@{}}Length\\ {[}m{]}\end{tabular}} & \multicolumn{2}{c}{FAST-LIO2 \cite{xu2022fast}}                                             & \multicolumn{2}{c}{LiLi-OM \cite{liliom}}                             & \multicolumn{2}{c}{KISS-ICP \cite{vizzo2023ral}}                               & \multicolumn{2}{c}{DMSA-LIO}                                              \\ \cmidrule(l){5-12} 
\multicolumn{1}{c}{}                         & \multicolumn{1}{c}{}                          &                                                                             &                                                                           & \multicolumn{1}{|c|}{rmse}           & \multicolumn{1}{c|}{max}            & \multicolumn{1}{c|}{rmse}  & \multicolumn{1}{c|}{max}   & \multicolumn{1}{c|}{rmse}   & \multicolumn{1}{c|}{max}     & \multicolumn{1}{c|}{rmse}           & \multicolumn{1}{c|}{max}            \\ \midrule
\multicolumn{1}{|l|}{Hilti-Oxford}           & \multicolumn{1}{l|}{exp04 Construction}       & \multicolumn{1}{c|}{125}                                                    & \multicolumn{1}{c|}{44.8}                                                 & \multicolumn{1}{c|}{0.024}          & \multicolumn{1}{c|}{0.091}          & \multicolumn{2}{c|}{N/A}                                & \multicolumn{1}{c|}{5.217}  & \multicolumn{1}{c|}{11.618}  & \multicolumn{1}{c|}{\textbf{0.022}} & \multicolumn{1}{c|}{\textbf{0.070}} \\ \midrule
\multicolumn{1}{|l|}{}                       & \multicolumn{1}{l|}{exp06 Construction}       & \multicolumn{1}{c|}{150}                                                    & \multicolumn{1}{c|}{54.2}                                                 & \multicolumn{1}{c|}{0.049}          & \multicolumn{1}{c|}{0.258}          & \multicolumn{2}{c|}{N/A}                                & \multicolumn{1}{c|}{69.961} & \multicolumn{1}{c|}{263.500} & \multicolumn{1}{c|}{\textbf{0.047}} & \multicolumn{1}{c|}{\textbf{0.228}} \\ \midrule
\multicolumn{1}{|l|}{}                       & \multicolumn{1}{l|}{exp14 Basement}           & \multicolumn{1}{c|}{74}                                                     & \multicolumn{1}{c|}{24.9}                                                 & \multicolumn{1}{c|}{0.200}          & \multicolumn{1}{c|}{0.369}          & \multicolumn{2}{c|}{N/A}                                & \multicolumn{2}{c|}{fails after 9\,s}                        & \multicolumn{1}{c|}{\textbf{0.068}} & \multicolumn{1}{c|}{\textbf{0.158}} \\ \midrule
\multicolumn{1}{|l|}{}                       & \multicolumn{1}{l|}{exp18 Corridor}           & \multicolumn{1}{c|}{109}                                                    & \multicolumn{1}{c|}{33.6}                                                 & \multicolumn{2}{c|}{fails after 87\,s}                                      & \multicolumn{2}{c|}{N/A}                                & \multicolumn{2}{c|}{fails after 5\,s}                        & \multicolumn{1}{c|}{\textbf{0.378}} & \multicolumn{1}{c|}{\textbf{0.732}} \\ \midrule
\multicolumn{1}{|l|}{Newer College}          & \multicolumn{1}{l|}{Parkland Mound}           & \multicolumn{1}{c|}{501}                                                    & \multicolumn{1}{c|}{443.2}                                                & \multicolumn{1}{c|}{31.242}         & \multicolumn{1}{c|}{45.524}         & \multicolumn{2}{c|}{fails after 52\,s}                    & \multicolumn{1}{c|}{0.254}  & \multicolumn{1}{c|}{1.384}   & \multicolumn{1}{c|}{\textbf{0.170}} & \multicolumn{1}{c|}{\textbf{1.365}} \\ \midrule
\multicolumn{1}{|l|}{}                       & \multicolumn{1}{l|}{Stairs}                   & \multicolumn{1}{c|}{118}                                                    & \multicolumn{1}{c|}{38.6}                                                 & \multicolumn{2}{c|}{fails after 38\,s}                                      & \multicolumn{2}{c|}{fails after 31\,s}                    & \multicolumn{2}{c|}{fails after 11\,s}                       & \multicolumn{1}{c|}{\textbf{0.047}} & \multicolumn{1}{c|}{\textbf{0.109}} \\ \midrule
\multicolumn{1}{|l|}{}                       & \multicolumn{1}{l|}{Cloister}                 & \multicolumn{1}{c|}{279}                                                    & \multicolumn{1}{c|}{288.5}                                                & \multicolumn{1}{c|}{0.077}          & \multicolumn{1}{c|}{0.315}          & \multicolumn{1}{c|}{0.766} & \multicolumn{1}{c|}{1.958} & \multicolumn{1}{c|}{0.632}  & \multicolumn{1}{c|}{3.338}   & \multicolumn{1}{c|}{\textbf{0.052}} & \multicolumn{1}{c|}{\textbf{0.144}} \\ \midrule
\multicolumn{1}{|l|}{Own data}               & \multicolumn{1}{l|}{Bicycle Street}           & \multicolumn{1}{c|}{1000}                                                   & \multicolumn{1}{c|}{1907.7}                                               & \multicolumn{1}{c|}{\textbf{5.870}} & \multicolumn{1}{c|}{\textbf{9.588}} & \multicolumn{2}{c|}{fails after 104\,s}                   & \multicolumn{1}{c|}{10.241} & \multicolumn{1}{c|}{17.248}  & \multicolumn{1}{c|}{15.986}         & \multicolumn{1}{c|}{24.060}         \\ \bottomrule
\end{tabular}}
\label{table:results}
\end{table*}

\subsubsection{Sliding Window DMSA Continuous Trajectory Optimization}\label{chap:sliding_window_opt}
In this step, a continuous trajectory of a sliding time window of duration $d_{window}$ is optimized.
The modeling of the trajectory was inspired by \cite{ramezani2022wildcat}. The trajectory is represented by a set of control poses uniformly distributed in time. Cubic Hermitian spline interpolation is performed between the control poses for the positions to obtain a temporal resolution of 1 ms. Between the orientations of the control poses spherical linear interpolation is used. The parameters of the control poses are then optimized using the DMSA algorithm with two extensions (see section \ref{chap:dmsa_core}). In addition to the downsampled points within the sliding time window, \textit{Static Points} are extracted from the map and added as fixed points within DMSA optimization. Furthermore, preintegrated IMU measurements (compare \cite{forster2016manifold}) between the control poses are added as a cost term at the Levenberg–Marquardt \cite{more2006levenberg} optimization step.

\subsubsection{Keyframe Selection and Representation}
New keyframes are selected based on the overlap with the map and the distance to the nearest keyframe (not the last). Using the distance to the nearest keyframe as a condition results in fewer redundancies in the global point cloud, what enables faster optimization. If a certain location is already densely mapped, the system implicitly switches to a localization mode, since no new keyframes are generated.
Downsampled point clouds from sliding time window optimization are used as keyframes. The planarity \cite{west2004context} is then calculated for each point of the downsampled point cloud as well as a normal vector based on their local neighborhood. The normal vectors are used once to remove non-visible points from the \textit{Static Points}, which are included in the sliding window optimization (compare section \ref{chap:sliding_window_opt}). In addition, the normal vectors are used in the subsequent keyframe optimization to split landmark point sets with opposite normal vectors.
For each keyframe, a gravity direction is estimated based on the estimated trajectory and IMU measurements within the optimized trajectory period. This gravity direction is included in the subsequent keyframe optimization. 
\subsubsection{DMSA Keyframe Optimization}
Adding a new keyframe triggers keyframe optimization. The first step is to analyze which part of the map should be optimized. To obtain a consistent map, the oldest keyframe is identified, whose distance to the current keyframe is smaller than a threshold value and whose overlap with the current keyframe is larger than a second threshold value. All keyframes that lie temporally between the identified oldest keyframe and the current keyframe are then optimized.
For keyframe optimization, the DMSA approach presented in section \ref{chap:dmsa_core} is used with two modifications. The first modification is that the planarity and the normal vectors of points within a landmark point set are analyzed. Here, planar point sets with opposing normal vectors are split into two sets based on the directions of the normal vectors, which are then processed as separate landmarks.
The second modification is that an additional error for the gravitation term is added during the optimization. This error term ensures that the gravitational directions estimated per keyframe point in the same direction in the global frame.

\section{Experiments}
To investigate the accuracy and robustness of our approach, we conducted experiments with sequences from the Hilti-Oxford Dataset \cite{9968057}, the Newer College Dataset \cite{ramezani2020newer} and own datasets. In selecting the sequences, the goal was to cover a wide range of conditions. Our selection includes construction sites, narrow staircases, long corridors, park landscapes and streets with dynamic objects. An important boundary condition for the selection was the presence of ground truth poses. While the Hilti-Oxford and the Newer College datasets contain groundtruth poses in centimeter range, we use a GNNS-based inertial navigation system as groundtruth for the evaluation of our own recording. Due to the length of the sequence, it is assumed here that the accuracy of the GNSS-based inertial navigation system is sufficient. In total, the sequences include data from three different LiDAR scanner types: Hesai PandarXT-32, Ouster OS1-64 and Ouster OS0-128.
\begin{figure}[h]
\begin{center}
		\includegraphics[width=.85\linewidth,angle=-2.5,origin=c,trim=0 30 0 55,clip]{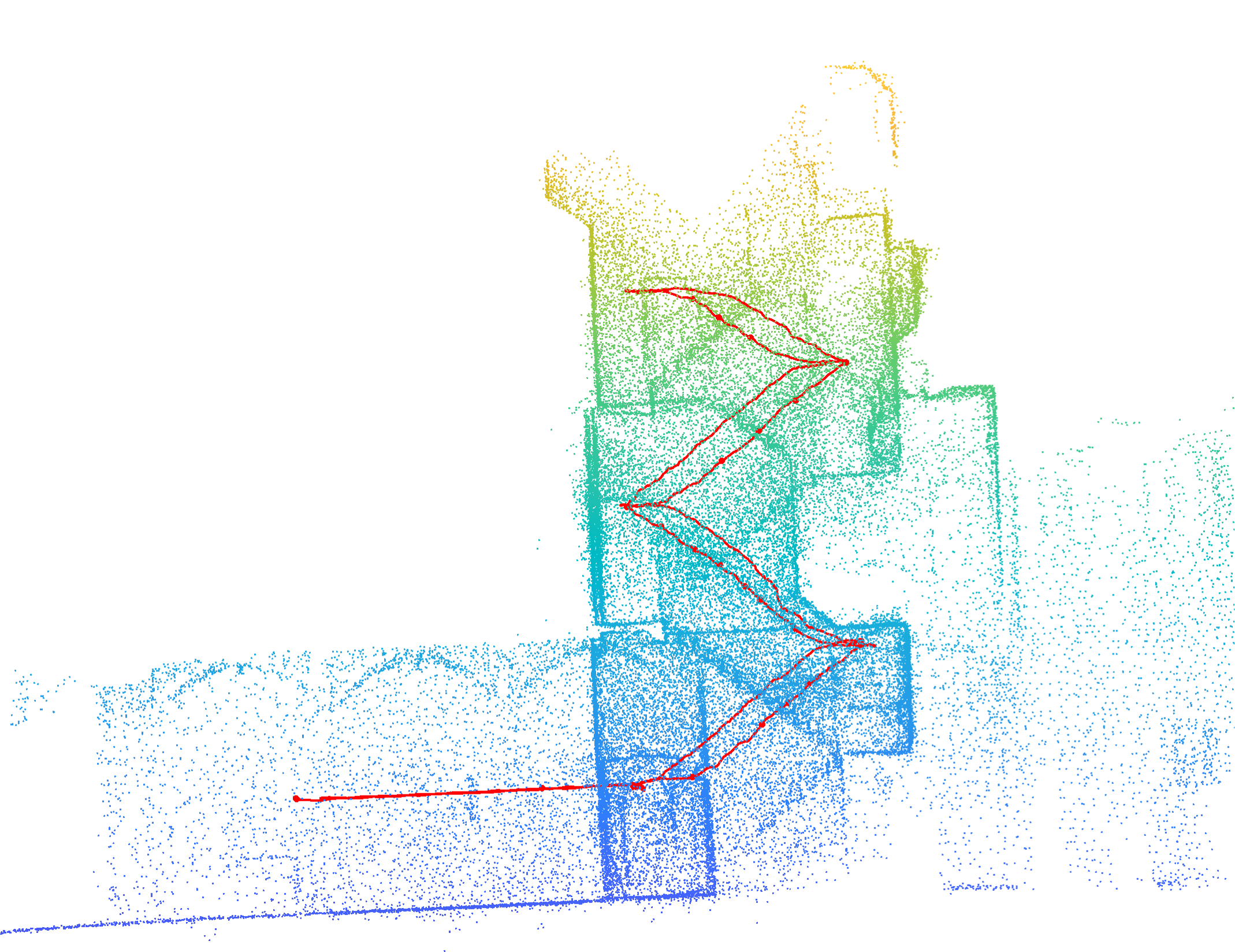}  
	\caption{Resulting sparse keyframe point cloud of sequence \textit{Stairs} of the Newer College Dataset \cite{ramezani2020newer}. The estimated trajectory is marked with a red line and keyframe positions are outlined with red dots.}
\label{fig:stairs}
\end{center}
\end{figure}
\begin{figure}
\begin{center}
		\includegraphics[width=0.8\columnwidth,trim=0 0 50 50,clip]{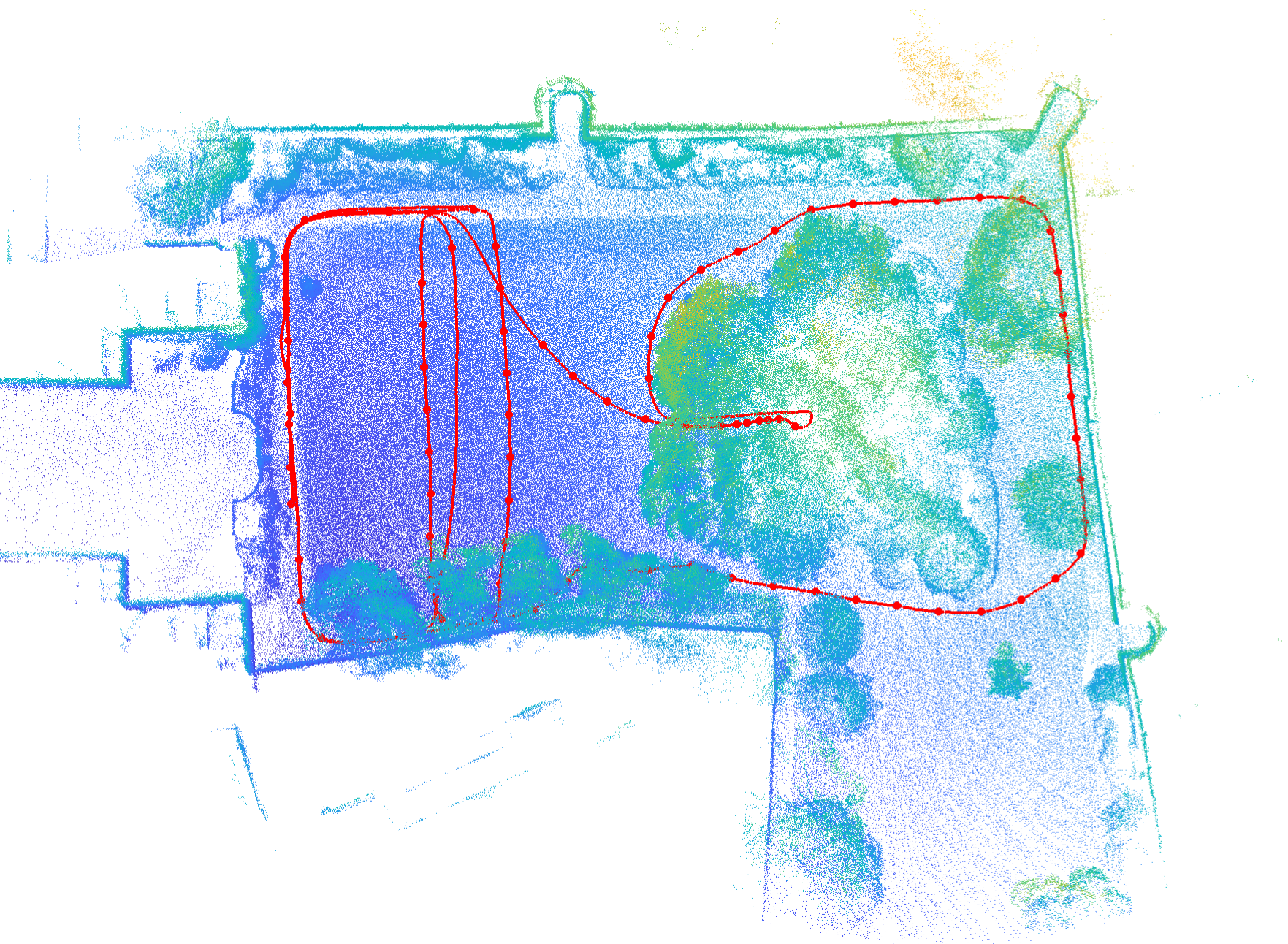}
	\caption{Resulting sparse keyframe point cloud of sequence \textit{Parkland Mound} of the Newer College Dataset \cite{ramezani2020newer}. The estimated trajectory is marked with a red line and keyframe positions are outlined with red dots.}
\label{fig:parkland_mound}
\end{center}
\end{figure}
\subsection{Selection of Comparing Approaches and Configuration}
For comparison, we also processed the sequences using other open source state of the art LiDAR-SLAM approaches. Here we selected FAST-LIO2 \cite{xu2022fast}, LiLi-OM \cite{liliom} and KISS-ICP \cite{vizzo2023ral}. While the first two combine LiDAR and IMU data like our approach, KISS-ICP \cite{vizzo2023ral} uses only the LiDAR data.
We did not perform parameter optimization for any of the selected methods. Instead, we used the default settings provided. For LiLi-OM \cite{liliom}, in addition to the rostopics and extrinsics, an adjustment had to be made in the source code to generate the correct LiDAR ring IDs during preprocessing. For the sensor setup of the Hilti-Oxford Dataset \cite{9968057}, we could not find a successful adaptation, so we noted corresponding values in Table \ref{table:results} with N/A.
In order to have a reasonable comparison, we also used for all sequences identical functional parameter settings in our approach.
\subsection{Processing Speed}
The processing time per sequence of DMSA-LIO varied between 1.5 and 4 times the recording time using a 12th Gen Intel Core i9-12900K. For indoor scenes the processing speed is significantly higher than for outdoor scenes. The current bottle neck is the numerical calculation of the Jacobian (see algorithm \ref{alg::core}).
\begin{figure}
\begin{center}
		\includegraphics[width=0.98\columnwidth,trim=0 30 50 70,clip]{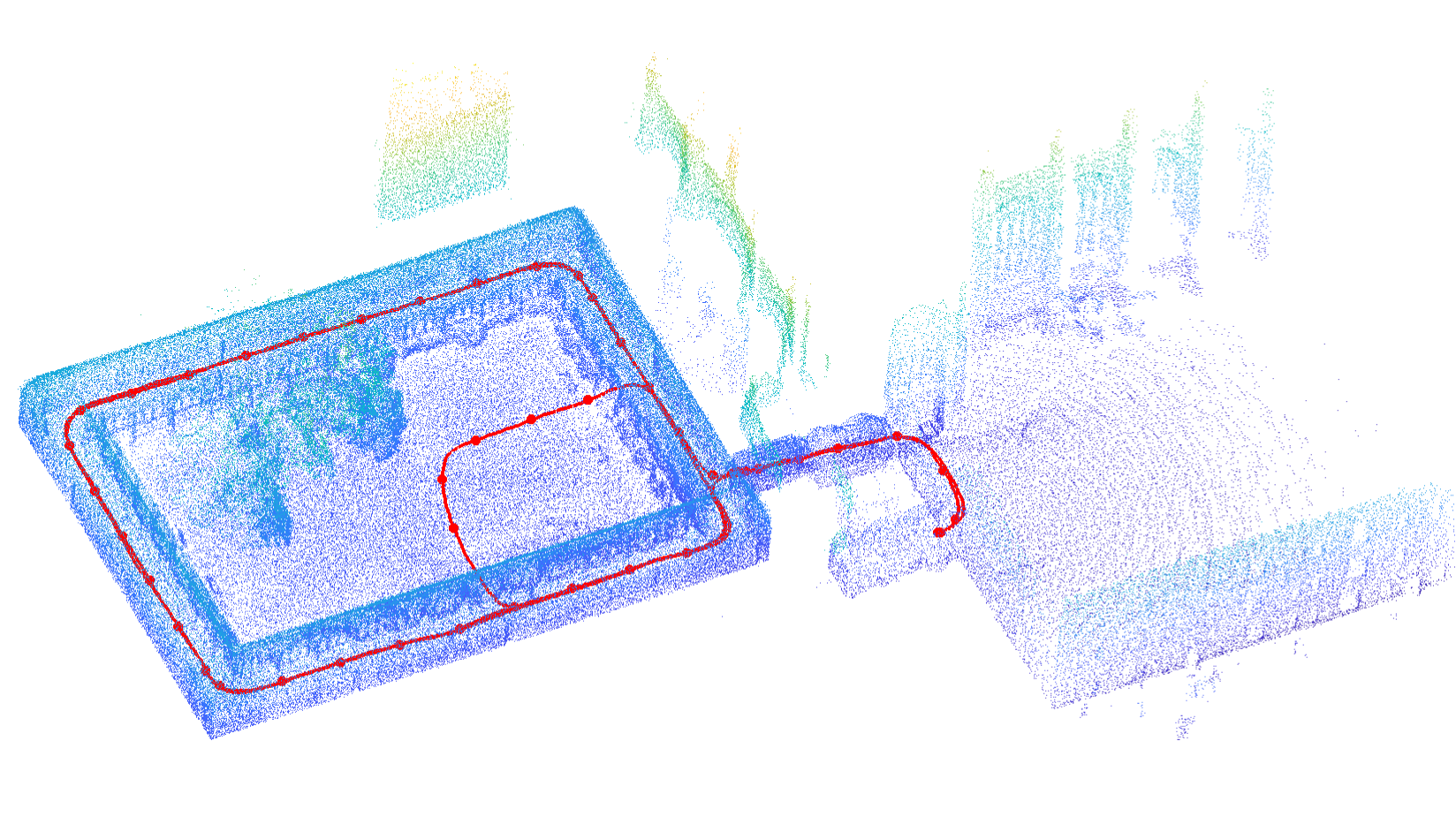}
	\caption{Resulting sparse keyframe point cloud of sequence \textit{Cloister} of the Newer College Dataset \cite{ramezani2020newer}. The estimated trajectory is marked with a red line and keyframe positions are outlined with red dots.}
\label{fig:cloister}
\end{center}
\end{figure}
\subsection{Results}
\begin{figure}[h]
\begin{center}
		\includegraphics[width=0.92\columnwidth,trim=15 35 13 75,clip]{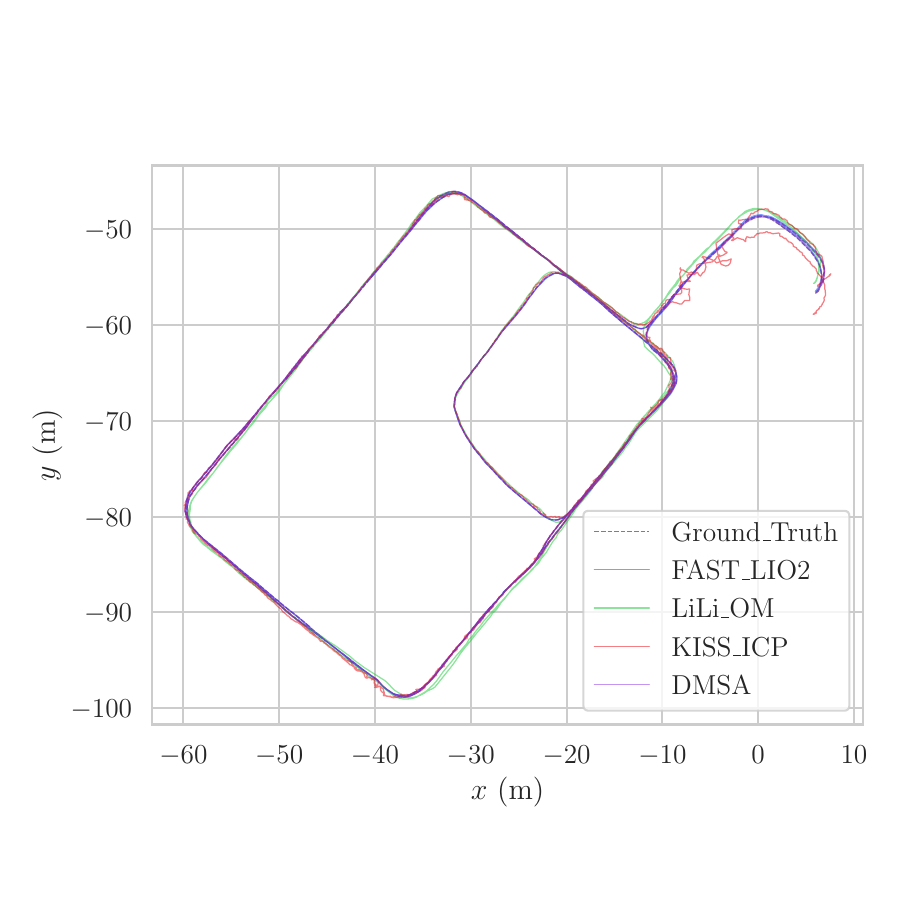}
	\caption{Estimated trajectories and ground truth for the \textit{Cloister} Sequence (with SE(3) Umeyama alignment) using \cite{grupp2017evo}.}
\label{fig:trajs_cloister}
\end{center}
\end{figure}

Fig. \ref{fig:stitch_with_circles} shows the keyframe point cloud and the estimated trajectory of our approach for the \textit{Bicycle Street} sequence. Despite the challenges due to dynamic objects and low structure locations, our method could be successfully applied here.
Fig. \ref{fig:stairs}, \ref{fig:parkland_mound}, and \ref{fig:cloister} show our results of the selected sequences from the Newer College Dataset \cite{ramezani2020newer}. Like the sequence \textit{exp14 Basement} from the Hilti-Oxford Dataset \cite{9968057} shown in Fig. \ref{fig:basement_bv}, these also include narrow spaces as a challenge.
A quantitative evaluation of the results is provided in table \ref{table:results}. Fig. \ref{fig:trajs_cloister} shows the estimated trajectories of the methods in the xy plane for the \textit{Cloister} sequence, where reasonable estimates were available from all approaches. Our method was able to provide a reasonable estimate for all selected sequences in the experiments. In contrast, FAST-LIO \cite{xu2022fast} failed in two sequences, LiLi-OM \cite{liliom} failed in three sequences and KISS-ICP \cite{vizzo2023ral} also failed in three sequences.
In addition, our approach was able to achieve the lowest rmse as well as lowest maximum error in seven from eight sequences.
\section{CONCLUSIONS}
This article presents a new method for dense multi scan adjustment. We also propose an extension of the method for application to continuous trajectories and show how this enables robust and accurate trajectory estimation. In the future, we aim to optimize the computational speed and further increase the accuracy. In addition, we also want to investigate a version of the method that works without an IMU.





\section*{ACKNOWLEDGMENT}

We would like to thank the German Federal Ministry for Economic Affairs and Climate Action for funding our project through the ZIM funding program (identifier KK5375601GR1).
We would also like to thank the company IGI mbH for the successful cooperation and the recording of sensor data.
\newpage
\bibliographystyle{IEEEtran}  
\bibliography{main}  
\end{document}